\documentclass[sigconf]{acmart}

\newcommand{\IMG}[1]{\mathbf{\mathcal{#1}}}

\newcommand{\RIG}{\mathcal{\mathbf{T}}}

\AtBeginDocument{%
  \providecommand\BibTeX{{%
    \normalfont B\kern-0.5em{\scshape i\kern-0.25em b}\kern-0.8em\TeX}}}

\acmConference[CVMP 2020]{the 16th ACM SIGGRAPH European Conference Visual Media Production}{Dec.\ 17--18}{London, UK}
\acmYear{2020}
\copyrightyear{2020}
\setcopyright{none}

\acmDOI{10.1145/1122445.1122456}

\acmBooktitle{Woodstock '18: ACM Symposium on Neural Gaze Detection, June 03--05, 2018, Woodstock, NY}
\acmPrice{15.00}
\acmISBN{978-1-4503-XXXX-X/18/06}

\usepackage{tabularx}
\usepackage{graphicx, caption}
\usepackage{cuted}
\graphicspath{{images/}} 

\begin{document}
\sloppy

\title{Neural Face Models for Example-Based Visual Speech Synthesis}

\author{Wolfgang Paier}
\email{wolfgang.paier@hhi.fraunhofer.de}
\affiliation{%
  \institution{Fraunhofer HHI}
}

\author{Anna Hilsmann}
\email{anna.hilsmann@hhi.fraunhofer.de}
\affiliation{%
  \institution{Fraunhofer HHI}
}

\author{Peter Eisert}
\email{peter.eisert@hhi.fraunhofer.de}
\affiliation{%
  \institution{Fraunhofer HHI}
}

\renewcommand{\shortauthors}{Trovato and Tobin, et al.}

\begin{abstract}
Creating realistic animations of human faces with computer graphic models is still a challenging task.
It is often solved either with tedious manual work or motion capture based techniques that require specialised and costly hardware.
Example based animation approaches circumvent these problems by re-using captured data of real people.
This data is split into short motion samples that can be looped or concatenated in order to create novel motion sequences.
The obvious advantages of this approach are the simplicity of use and the high realism, since the data exhibits only real deformations.
Rather than tuning weights of a complex face rig, the animation task is performed on a higher level by arranging typical motion samples in a way such that the desired facial performance is achieved.
Two difficulties with example based approaches, however, are high memory requirements as well as the creation of artefact-free and realistic transitions between motion samples.
We solve these problems by combining the realism and simplicity of example-based animations with the advantages of neural face models. 
Our neural face model is capable of synthesising high quality 3D face geometry and texture according to a compact latent parameter vector.
This latent representation reduces memory requirements by a factor of 100 and helps creating seamless transitions between concatenated motion samples.
In this paper, we present a marker-less approach for facial motion capture based on multi-view video.
Based on the captured data, we learn a neural representation of facial expressions, which is used to seamlessly concatenate facial performances during the animation procedure. We demonstrate the effectiveness of our approach by synthesising mouthings for Swiss-German sign language based on viseme query sequences.
\end{abstract}



\begin{teaserfigure}
  \includegraphics[width=\textwidth]{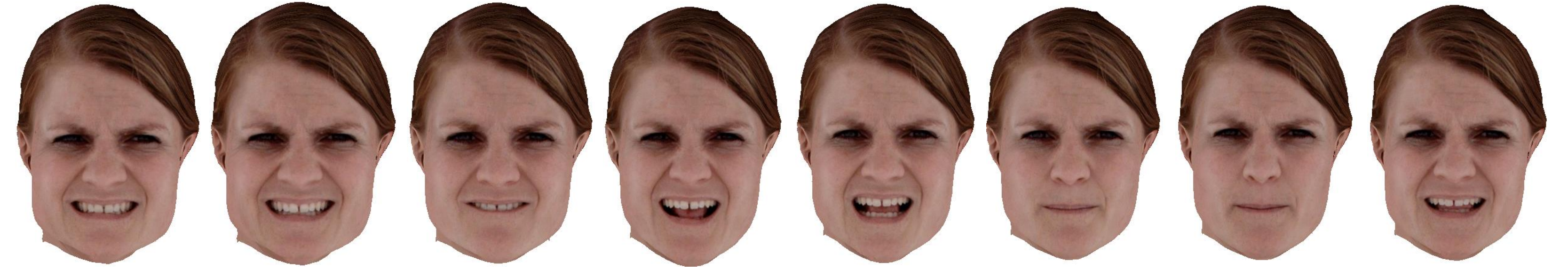}
  \caption{Selected frames of a visual speech sequence when pronouncing CVMP.}
  \label{fig:teaser}
\end{teaserfigure}

\maketitle

\section{Introduction}
Animation and rendering of realistic human faces is one of the most challenging tasks in computer graphics.
This is due to the fact that the human face is a complex object composed of bones, muscles, several tissue layers and skin, which results in complex deformations that are difficult to represent with a simple 3D mesh.
Reflective properties tend to be complex as well. They vary from diffuse to highly specular and include even subsurface scattering effects.
Furthermore, areas like the mouth and the oral cavity pose another problem during the automatic creation of animatable 3D face models:
reflective properties as well as occlusions/diss-occlusions (oral cavity) tend to produce artefacts, which makes a detailed and holistic 3D representation (such as a triangle mesh) rather difficult to create.
For realistic animations of human faces, especially during speech synthesis, the dynamics (e.g. the deformation and the speed, at which the face changes from one expression to another) plays an important role as well.
Inappropriate timings of visemes or transitions between visemes may cause an unnatural impression of speech and even degrade the comprehensibility of the virtual human. Moreover, humans are very good at interpreting faces, such that small deviations from the expected appearance or behaviour are perceived as wrong or unnatural \cite{Seyama2007}, which might decrease the acceptance of potential users.
Therefore, we propose a novel hybrid animation approach for realistic visual speech synthesis with animatable 3D face models.
Our system addresses the two previously outlined problems, which may occur during automatic creation and animation of 3D human face models.
We ensure realistic appearance and deformation capabilities by employing a neural face model similar to \cite{paier2020} that adapts 3D geometry and texture simultaneously according to a compact parameter vector.
This enables us to capture detailed facial expressions and deformation using high resolution textures and a coarse proxy geometry.
While the geometry model captures rigid motion and large scale deformation of the face, we use the neural texture representation to capture fine movements, details like wrinkles or changes in the appearance of skin.
In order to ensure realistic dynamics, we base our animation method on the idea of example-based animation, e.g.~motion graphs \cite{Kovar2002}.
We create a set of short motion samples that represent typical or atomic motions. Selected transition points between samples allow looping and/or concatenating them, which enables the user to create novel performances that have not been captured before.
Since these atomic motion snippets are created by capturing a real human, the  visual quality is very high as they simply reproduce the original movements of the captured person.
The challenge, however, is the creation of natural and seamless sequence of motion samples that does not show artefacts at transitions between consecutive snippets.
This includes static artefacts, e.g. differences in geometry/texture at the transition point but also dynamic artefacts like sudden changes in motion.
To perform visual speech synthesis, we capture an actress during speech with video cameras and create a database of short motion samples
of her mouth (dynamic visemes). Visemes can be interpreted as the visual equivalent of phonemes and represent atomic visual units of speech.
This includes the shape of the mouth, lips, tongue and the visibility of teeth.
By concatenating dynamic visemes, we are able to synthesise new performances that have not been captured before.
The input for our system consists only of a viseme query sequence. Using graph-cuts, we choose an optimal viseme sequence from our database.
Our method considers the concrete viseme label, effects like co-articulation and minimises static as well as dynamic artefacts at transition points. 
In order to create truly seamless and artefact free transitions, we rely on the interpolative capabilities of our neural face model, which enables us to match mouth expressions at transition points by performing interpolation in latent space.
The remainder of this paper is organised as follows: in section \ref{sec:sota}, we present an overview of existing technologies and methods that are related to facial animation and facial modelling as well as visual speech synthesis.
Section \ref{sec:markperfcap} presents the database creation process, which includes the marker-less facial performance capture and the training of the neural face model.
Section \ref{sec:visspeechsyn}, contains a detailed description of the proposed visual speech synthesis approach.
In the last two sections, we present and discuss our results.

\section{Related Work}\label{sec:sota}

\subsubsection*{Modelling of Facial Expressions.}
Well known 3D face representations, like \cite{Cootes98, eigi98, Blanz1999, Vlasic2005, Cao2014}, were based on linear basis shapes and even though their expressive power is limited, they became popular due to their robustness and simplicity of use.
In order to deal with the limitations, researchers proposed different strategies to increase the quality of captured facial expressions.
For example, Li et al. \cite{Li2013} estimates corrective blend-shapes on the fly.
Another example can be found in \cite{Garrido2013}, where Garrido et al. fit blend-shapes to a detailed 3D scan using sparse facial features and optical flow.
The resulting personalised face model is then used to synthesise plausible mouth animations of an actor in a video according to a new target audio track.
A more sophisticated facial retargeting system was proposed by Thies et al. \cite{Thies2015} who implemented a linear model that can represent expressions, identity, texture and light.
Their model it is able to capture a wide range of facial expressions of different people under varying light conditions.
However, a major drawback of these simple linear models is that they cannot properly capture complex areas like the mouth cavity as they lack the expressive power to capture complex deformations and occlusions/dis-occlusions that occur around eyes, mouth and in the oral cavity.
As a result, approaches that are based on linear face representations \cite{Garrido2013, Thies2015} only, use either 'hand crafted' solutions to visualise complex areas (e.g. oral cavity) or simply ignore them.
Other works like \cite{Paier2017, FechtelerPaierHilsmannEisertICIP2016, Paier2015, Casas2014, FechtelerPaierEisertICIP2014, Dale2011, Lipski11, Kilner06, Borshukov2006, Carranza2003a}, try to circumvent these limitations
by employing image-based rendering approaches, where the facial performance is captured in geometry and texture space.
In \cite{Dale2011}, Dale et al. present a low cost system for facial performance transfer. Using a multilinear face model, they are able to track an actor's facial geometry in 2D video, which enables them to transfer facial performances between different persons via image-based rendering. Similar techniques were used in \cite{FechtelerPaierEisertICIP2014}, \cite{FechtelerPaierHilsmannEisertICIP2016} or \cite {Thies2015}.
Another image based animation strategy was proposed in \cite{Casas2014,Paier2017,Paier2015}, where low resolution 3D meshes were combined with highly detailed dynamic textures that allow generating high quality renderings.
Based on previously captured video and geometry, they create a database of atomic geometry and texture sequences that can be re-arranged to synthesise novel performances.
While Casas et al. use pre-computed motion-graphs to find suitable transitions points, Paier et al. proposed a spatio-temporal blending to generate seamless transitions between concatenated sequences.
The big advantage of these approaches is that they make direct use of captured video data to synthesise new video sequences, which results in high quality renderings.
Geometry is only used as a proxy to explain rigid motion, large scale deformation and perspective distortion.
However, the high visual quality comes at the cost of high memory requirements during rendering, since textures have to be stored for each captured frame in the database.
With the advent of deep neural networks, more powerful generative models have been developed. Especially, variational auto encoders \cite{kingma2013autoencoding} VAE and generative adversarial networks \cite{Goodfellow2014} (GAN) gain popularity, since they are powerful enough to represent large distributions of high dimensional data in high quality. 
For example Nagano et al. \cite{nagano2018} trained a generative adversarial network that is capable of synthesising facial key expressions according to a single input image.
With these key expression images and a set of corresponding blend-shapes, they are able to derive a personalised blend-shape model with key textures. During rendering, these key textures are blended linearly to achieve the desired appearance.
Another learning based approach is proposed by Pumarola et al. \cite{pumarola2018} who present a GAN that is able to generate facial expressions based on an action unit (AU) coding scheme.
The action unit weights define a continuous representation of the facial expressions. The loss function is defined by a critic network that evaluates the realism of generated images and fulfilment of the expression conditioning.
While these approaches produce impressive results, they are still limited by the fact that the underlying representation is restricted to blend-shape/AU weights.
A different method was presented by Lombardi et al. \cite{Lombardi2018, Slossberg2018, paier2020}. They train a deep neural network, which represents geometry and texture with a single latent feature vector that allows reconstructing facial geometry as well as texture. In contrast to previous approaches, this allows capturing all facial expressions in high quality, since the representation is not tied to a simple linear model (e.g. FACS or linear blend-shapes). 

\subsubsection*{Visual Speech Synthesis.} Visual speech synthesis refers to the process of performing mouth animation according to a speech signal.
This can be accomplished, for example, by performing a direct mapping from audio to animation parameters using a regression function \cite{thies2020, shimba2015, hou2016b, suwajanakorn2017, karras2017, zhou2018, cudeiro2019, hussenabdelaziz2019}.
For example Shimba et al. \cite{shimba2015} formulate the task as a sequence-to-sequence mapping with audio as input and video as output
They represent the audio signal via mel-frequency cepstral coefficients (MFCC), AMFCC and A2 MFCC. The visual output signal is represented by an active appearance model (AAM) on the entire face region.
Their regression model is composed of long short-term memory (LSTM) units that learn a direct mapping from audio to visual features.
A similar approach is proposed by Zhou et al. \cite{zhou2018}. While they rely on similar audio features as \cite{shimba2015}, they use an animator centred face rig to represent facial expressions.
This allows creating sequences of visual speech that can be edited and fine-tuned by a human animator. Since the training data for the face rig needs to be created manually, they had to deal with the problem of training a deep neural regression model with a small number of training samples. To achieve this, they implemented a two-phase training strategy.
In a pre-training step, they use large existing databases to learn a latent representation of the audio signal by predicting facial landmarks and phoneme groups.
In the fine-tuning step, they extend their network to predict the target animation parameters from audio features, landmarks and phoneme groups.
While most recent regression based approaches implement recurrent neural networks, \cite{karras2017, cudeiro2019} proposed convolutional neural networks to animate a 3D face geometry model. Karras et al. present in \cite{karras2017} a convolutional neural network architecture that takes waveforms as input and outputs vertex positions.
Additionally, they account for different speaking styles (i.e. due to different speakers or emotions) by learning an emotion database that stores a latent emotion vector for each training sample. 
This allows the network to store appearance information, which cannot be inferred from audio alone in the emotion database. The emotion vector is regularised by reducing its length to ensure that the network learns a valid mapping from audio to geometry.

Another typical way of performing visual speech synthesis is based on the concatenation and interpolation of already existing data \cite{bregler1997, cosatto2000, ezzat2002, taylor2012}.
For example, Taylor et al. \cite{taylor2012} extract prototypes of visemes based on large audio-visual corpus. They represent mouth expressions with an active appearance model (AAM). Based on the gradient of AAM parameters, they extract gesture boundaries. Using a Gaussian mixture model (GMM), they are able to identify 150 dynamic visemes. It is worth noting, that they choose a rather large number of visemes (compared to phonemes), because they intent to explain co-articulation effects as well.

In this paper, we propose the combination of neural face models to represent facial expressions with a concatenative approach for visual speech synthesis. Similar to \cite{paier2020}, we use a VAE based architecture with an additional adversarial loss.
In contrast to regular VAE based approaches \cite{Lombardi2018}, the GAN-loss forces the generator network to produce more realistic textures with more details.
As the input-signal for our system consists of viseme labels (phoneme groups) only, we choose a concatenative approach, since working with original visual speech samples avoids under-articulation and yields a richer and clearer speech animation. Additionally, by employing the neural face model, we are even able to solve two typical problems of
example-based animation approaches: artefacts during interpolation of facial expressions and high memory requirements for storing the captured data in memory.




\section{Markerless Performance Capture}\label{sec:markperfcap}
In order to perform visual speech synthesis based on example data, we need to create a database with samples of typical facial motion during speech.
Since our face representation builds on dynamic textures as well as a dynamic proxy geometry, we rely on a marker-less approach for facial performance capture like \cite{paier2020}.
During performance capture, we use only an approximate geometry model of the actress face to explain rigid motion and large scale deformations of mouth and jaw.
Fine deformations like wrinkles and small motions are captured in texture space.
This approach has two big advantages:
first, we do not need to create a very accurate face geometry model, which means that even linear face models could be re-used.
Second, by capturing all remaining information (i.e. that cannot be explained with the geometry model) in texture space, we do not lose any details of the facial expression during performance capture.

\subsection{3D Pose and Shape Estimation}
In order to estimate the 3D head pose as well as the approximate geometric deformation, we use a PCA model $\mathcal{\mathbf{B}}$ of the actress face. We optimise the following parameters during mesh registration: rigid motion $\RIG$ is represented by a six dimensional parameter vector $\mathbf{t}=[tx ,ty, tz, rx, ry, rz]^T$ that contains three parameters for translation $(tx, ty, tz)$ and three parameters for rotation $(rx, ry, rz)$.
The weight vector $\mathbf{b} = [b_0,...b_{14}]^T$ is a fifteen dimensional column vector containing the shape weights and $\mathcal{\mathbf{B}}$  represents the linear basis of facial expressions, which consists of a three dimensional vector for each vertex in the template mesh with $n$ vertices.

    \begin{equation}
    \mathbf{x}_{\mathbf{t},\mathbf{b}} = \RIG\big(\mathcal{\mathbf{x_0}} + \mathcal{\mathbf{B}} \mathbf{b}\big)
    \label{eq:vertexmodel}
    \end{equation}

The column vector $\mathbf{x}_{\mathbf{t},\mathbf{b}}=[x,y,z,...,x_n,y_n,z_n]^T$ contains $x, y$ and $z$ coordinates of all vertices of the deformed mesh. $\mathcal{\mathbf{x_0}}$ corresponds to the mean shape of the mesh template and $\mathcal{\mathbf{T}}$ represents the rigid transform that is applied as a last step on all deformed vertices.
$\mathbf{x}_i$ refers to the 3D position $[x_i,y_i,z_i]^T$ of the i-th transformed vertex.
The input data for our facial performance capture approach consists of calibrated and synchronised multi-view video streams and facial landmarks.
We asked our actress to start at least one take with a neutral facial expression, which is used as a reference frame during tracking. 
This eases the performance capture process, while not posing a big limitation on our method.
For automatic facial landmark detection we use the method proposed in \cite{Kazemi2014}.
As an initial step during performance capture, we register the linear face model to one multi-view video frame where the actress shows a neutral face.
This frame acts as a reference for all subsequent pose and shape estimations.
During this registration step, we minimize the distance between the detected 2D landmark positions $\mathbf{m}_{c,i}$ and $\mathbf{\hat{m}}_{c,i}$, the corresponding 3D location on the face geometry model projected on the image plane of camera $c$.

    \begin{equation}
    \mathcal{E}_{lm}(\mathbf{t},\mathbf{b}) = \sum_{c}^C{\sum_{i} \big|\mathbf{m}_{c,i} - \mathbf{\hat{m}}_{c,i}\big|^2},
    \label{eq:lmerror}
    \end{equation}

We use the result of this initial registration as a reference for the whole performance capture process.
In order to estimate the 3D head pose as well as blend-shape weights for all captured frames, we minimise the landmark error (\ref{eq:lmerror}) as well as
intensity difference (\ref{eq:imgerror}) between captured images and the rendered face model.

    \begin{equation}
    \mathcal{E}_{img}(\mathbf{t},\mathbf{b}) = \sum_{c}^C{\sum_{p \in I} \big|\IMG{J}_c(p) - \IMG{I}_{\mathbf{t},\mathbf{b}}(p)\big|^2},
    \label{eq:imgerror}
    \end{equation}

For each camera $c$, we render the tracked face model and use the captured reference frame of camera $c$ as texture.
We treat this registration process as a non-linear optimization task. In order to make the estimation of shape weights $\mathbf{b}$ more stable, we include an $L2$ regularization term in the objective function  (\ref{eq:trackingerror}).

    \begin{equation}
    \mathcal{E}_{tracking}(\mathbf{t},\mathbf{b}) = \mathcal{E}_{img} + \lambda \mathcal{E}_{lm} + \gamma |\mathbf{b}|^2
    \label{eq:trackingerror}
    \end{equation}

The scalars $\lambda$ and $\gamma$ are weight factors to control the influence of the landmark data term (\ref{eq:lmerror}) and shape regularization.
We minimise the tracking objective (\ref{eq:trackingerror}) using the Gauss-Newton method.

\begin{figure}[!ht]
\includegraphics[width=0.9\columnwidth]{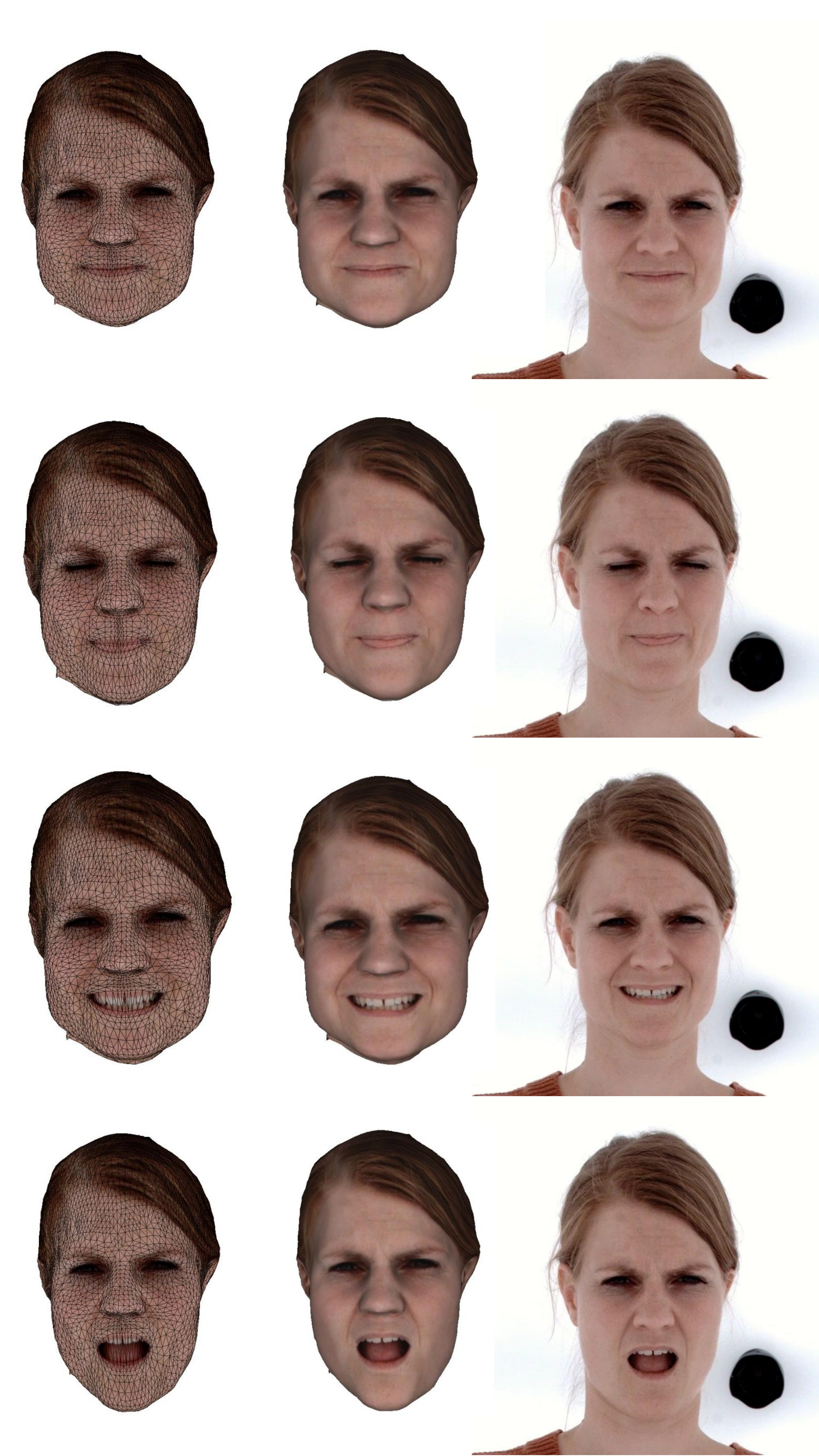}
\protect\caption{Results of the facial performance capture process. Each rows shows a different facial expression. The left and middle column show the textured face model (left: wireframe, middle: with directional light). The right column shows the original video frames. It is worth noting that complex areas like mouth and eyes are properly captured and can be reconstructed in high quality. The wireframe renderings in the left column, for example show the oral cavity, which is not represented in geometry at all. Even though tongue, teeth and gums are captured only in texture space, the rendered model looks realistic. }
\label{fig:perfcap}
\end{figure}

\subsection{Texture Extraction}
Knowing the accurate pose and approximate face shape for each frame in the captured multi-view videos allows extracting a dynamic texture sequence for each multiview video.
These dynamic texture sequences capture all information that cannot be represented by geometry. For example, fine motions and deformations (e.g. micro-movements, wrinkles), changes in texture and occlusions/dis-occlusions (e.g. eyes and oral cavity).
Together, geometry and texture represent almost the full appearance of the actress face in each frame.
In order to generate high quality dynamic textures, we rely on a graph-cut based approach that implements a temporal consistency constraint \cite{Paier2015}.
This method simultaneously optimises three data terms (\ref{eq:TexSeamOptTemporal}) to create a visually pleasing (i.e. no spatial or temporal artefacts) sequence of textures from each multi-view video:

\begin{equation}
\begin{split}
\mathcal{E}_{\mbox{tex}}(C)=&\sum_{t}^{T}\sum_{i}^{N}\mathcal{D}(f_{i}^{t},c_{i}^{t})\\
													&+\lambda\sum_{i,j\in\mathcal{N}}\mathcal{V}_{i,j}(c_{i}^{t},c_{j}^{t})\\
													&+\eta\mathcal{T}(c_{i}^{t},c_{i}^{t-1})
\label{eq:TexSeamOptTemporal}
\end{split}
\end{equation}

$C$ denotes the set of source camera ids for all mesh triangles. The
first term $\mathcal{D}(f_{i},c_{i})$ defines a measure for the visual quality
of a triangle $f_{i}$ textured by camera $c_{i}$. It uses the heuristic
$\mathcal{W}(f_{i},c_{i})$, which is the area of $f_{i}$ projected
on the image plane of camera $c_{i}$ relative to the sum of $area(f_{i},c_{i})$
over all possible $c_{i}$ to ease the choice of the weighting factors
$\eta$ and $\lambda$:
\begin{equation}
\mathcal{D}(f_{i},c_{i})=\begin{cases}
1-\mathcal{W}(f_{i},c_{i}) & f_{i}\text{ is visible}\\
\infty & f_{i}\text{ is occluded}
\end{cases}\label{eq:regional_term}
\end{equation}
\begin{equation}
\mathcal{W}(f_{i},c_{i})=\frac{area(f_{i},c_{i})}{\underset{c_{j}}{\sum}area(f_{i},c_{j})}\label{eq:WeightFunc}
\end{equation}
$\mathcal{V}_{i,j}(c_{i},c_{j})$ represents a spatial smoothness constraint, which relates to the sum of colour differences along the common edge
$e_{i,j}$ of two triangles $f_{i}$ and $f_{j}$ that are textured from different cameras $c_{i}$ and $c_{j}$.
\begin{equation}
\mathcal{V}_{i,j}(c_{i},c_{j})=\begin{cases}
0 & c_{i}=c_{j}\\
\Pi_{e_{i,j}} & c_{i}\neq c_{j}
\end{cases}\label{eq:boundary_term}
\end{equation}
\begin{equation}
\Pi_{e_{i,j}}=\int_{e_{i,j}}\left\Vert I_{c_{i}}(x)-I_{c_{j}}(x)\right\Vert dx\label{eq:boundary_term_cost}
\end{equation}
The last  term $\mathcal{T}(c_{i},c_{j})$ in (\ref{eq:TexSeamOptTemporal}) ensures temporal smoothness by
penalising changes of the source camera $c_{i}$ of a triangle $f_{i}$ between consecutive time steps. 
Without such a term, the extracted dynamic textures are not temporally consistent, i.e.~the source camera
of a triangle can change arbitrarily between two consecutive texture frames, which introduces flickering artefacts in the resulting texture sequence.

\begin{equation}
\mathcal{T}(c_{i}^{t},c_{i}^{t-t})=\begin{cases}
0 & c_{i}^{t}=c_{i}^{t-t}\\
1 & c_{i}^{t}\neq c_{i}^{t-t}
\end{cases}\label{eq:boundary_term_temp}
\end{equation}

The selection of source cameras $C$ for all mesh-triangles in all frames is solved simultaneously using the alpha expansion algorithm \cite{Boykov2001}. 
In a last step, we assemble the reconstructed textures by sampling texel colors from the optimal source cameras. 
Seams between triangles are concealed using gradient domain blending \cite{Perez2003}.

\subsection{Neural Face Representation}

\begin{figure}[!ht]
\includegraphics[width=\columnwidth]{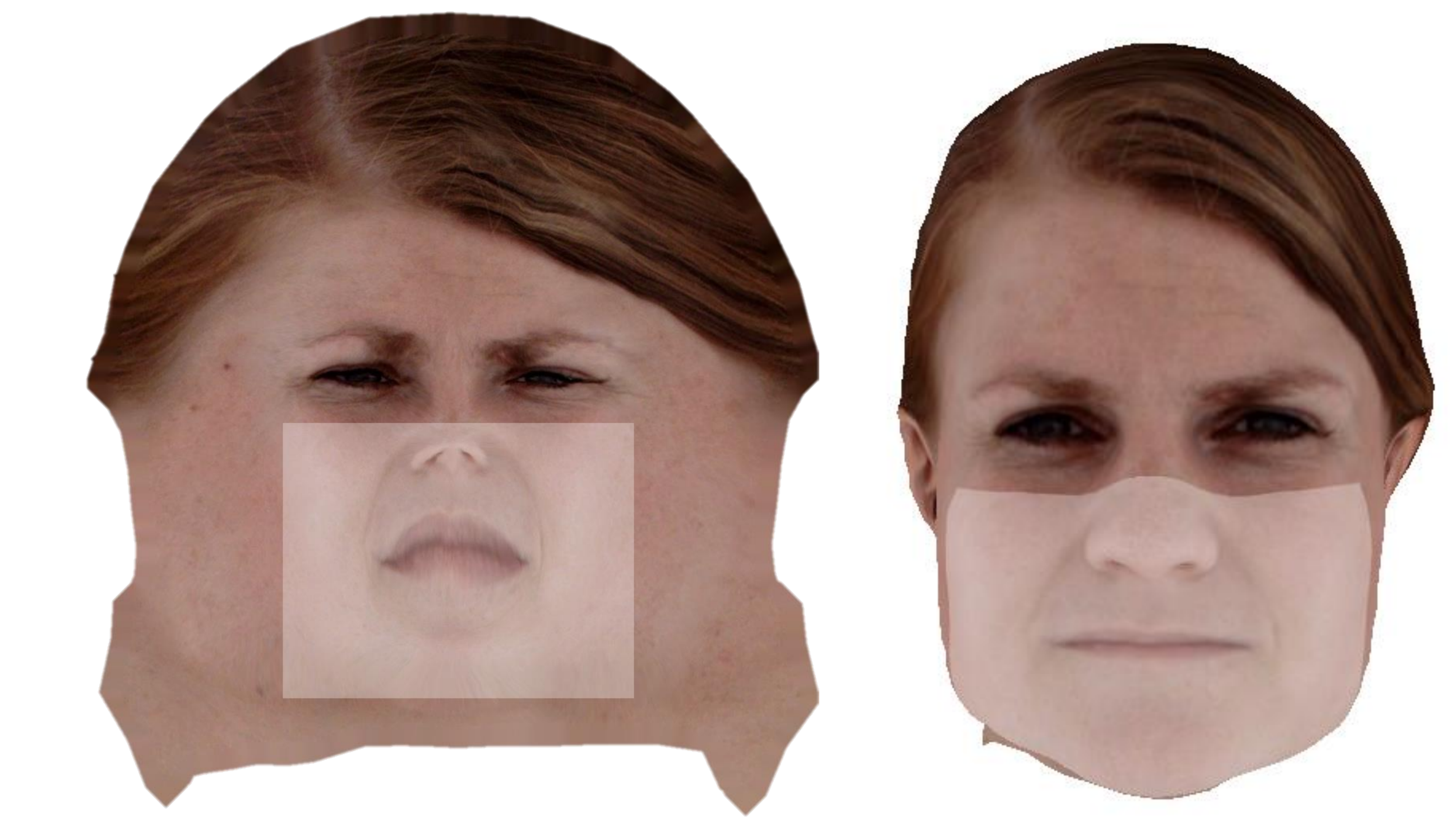}
\protect\caption{We select the modified mouth area in texture space by defining a rectangular region of interest.
Similarly, we select the corresponding vertices of interest by comparing the texture coordinate of each vertex with the region of interest in texture space.
This creates a consistent region interest for the complete face model (geometry and texture). }
\label{fig:mouthroi}
\end{figure}

By running our facial performance capture algorithm on the recorded data, we transform the multi-view face video into a sequence of consistent face meshes with a separate texture for each frame.
Since all meshes have the same triangle definition, uv-coordinates and an equal number of vertices, the uv-mapping is constant as well.
This consistent face representation supports the definition of a local face animation region. For visual speech synthesis, we only concentrate on the mouth area.
We select the mouth area by defining a region of interest in texture space, figure \ref{fig:mouthroi}. This rectangular texture area as well as all vertices that belong to this area are represented by our neural model as detailed below.

\begin{figure*}[!h]
\includegraphics[width=0.95\textwidth]{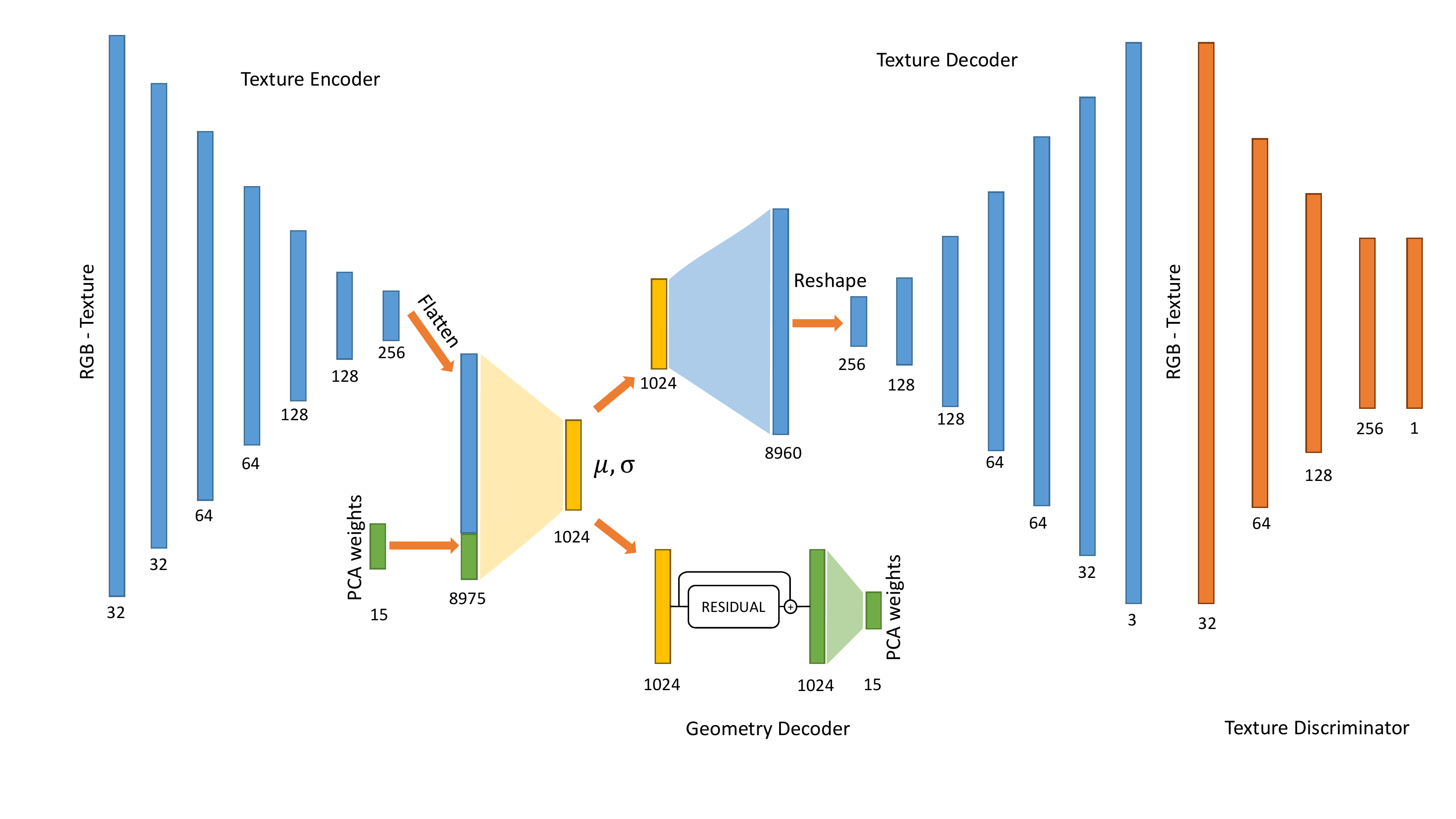}
\protect\caption{This figure shows the used network architecture of the neural face model. The network consists of five parts: a convolutional texture encoder/decoder (blue), a geometry decoder (green), 
a fully-connected bottleneck (yellow) that combines information of texture and geometry into a latent code vector ($\mu$), and deviation ($\sigma$). And a texture discriminator network (orange) that classifies textures as real or as synthetic.}
\label{fig:arch}
\end{figure*}

The reason for using this neural representation is two-fold: first, the parametric representation supports realistic interpolation between different expressions.
This is especially useful, since our 3D face model acts only as a geometric proxy.
This means that effects like small deformations or occlusions/dis-occlusions have to be captured in texture space (e.g.~teeth and tongue becomes visible when the mouth is opened).
A linear texture model (e.g. PCA) as well as simple non-linear models (GPLVM) are usually not capable of performing this kind of interpolations in high quality as intermediate textures are too blurry or contain ghosting artefacts \cite{paier2020}.
The second advantage of using a neural representation is low memory requirements for storing captured data.
For example, the raw texture data of the mouth region consists of 480x370 RGB pixels and the PCA weights to describe the geometry (i.e. 15 float values), which results in approximately 500kB per frame.
The latent representation of mouth texture and geometry consists of 1024 float values, which results in approximately 4kB per frame.
Using the latent representation we can store approximately 11 hours of sample data in 4GB RAM. Compared to approximately 5.5 minutes of samples using raw data.

Figure \ref{fig:arch} shows the architecture of our neural model based on a deep VAE \cite{kingma2013autoencoding}.
We feed the estimated PCA weights from performance capture as well as the mouth textures into the auto-encoder.
The texture-encoder transforms the mouth texture to its latent representation. This latent texture feature vector is concatenated with the geometry PCA weights and transformed to a 1024 dimensional latent representation of the mouth expression.
The latent expression vector is then processed by the texture decoder and geometry decoder separately.
Each texture processing layer in the encoder and decoder performs a double-convolution (two consecutive blocks of zero-padding, conv3x3, batch-norm, leaky-ReLU).
The encoder performs a 2x2 max-pooling at the end to of each double-convolution to reduce the resolution. The decoder uses a transposed 2x2-convolution with stride 2 to double the resolution at the beginning of each decoding block.
As suggested by \cite{Lombardi2018}, we add a spatially varying bias layer at the end of each texture decoding block to improve convergence and texture quality.

Our discriminator network is based on the PatchGAN structure as described in \cite{Isola2016}. It consists of 4 convolutional blocks. Each block consists of a 2d convolution with stride 2, batch-norm and leaky-ReLU layer.
The final layer is a 1x1 convolution that outputs a single channel feature-map, where a value close to zero means that the discriminator believes that this image region is fake, while a value close to one signals that the image patch is real.
The advantage of using smaller patches is that the discriminator classifies based on smaller features and details, which forces the texture generator to create a sharper texture with more details.
The geometry decoder consists of a simple residual block: two fully connected layers (linear layer, batch-norm, leaky-ReLU) compute a residual that is added to the latent expression vector.
A linear output layer transforms the updated expression vector to the target PCA weights.
The full objective function consists of the absolute difference between the predicted texture and the target texture, the adversarial loss of the PatchGAN and the mean squared error between the predicted PCA weights and the target PCA weights.

We use a batch-size of 4, a leakiness of 0.01 for all ReLUs in our network and use the Adam optimiser with default parameters to train our network for 100 epochs with an initial learning rate of 0.00001 and exponential learning rate scheduling ($\gamma=0.95$).

\section{Visual Speech Synthesis}\label{sec:visspeechsyn}
Based on the extracted facial performance capture data, we create a database of annotated visual speech samples.
Each sample is defined by its latent representation (i.e. a sequence of latent feature vectors) and a viseme label.
For our experiments, we create a visual speech dataset that consists of five takes and has an overall length of approximate 7.2 minutes.
We captured several single words as well as phrases related to weather and a full weather forecast in German language.
Our viseme dictionary is based on the research in \cite{Elliott2013}, but in order to faithfully reproduce the visual speech style of our actress, we extended the initial set with three more visemes. The full viseme dictionary consists of 13 visemes that are listed in table \ref{tab:visemes}.
In order to generate more accurate results, we annotate the captured data manually.
\begin{table}
  \caption{Viseme Dictionary}
  \label{tab:visemes}
  \begin{tabular}{ccl}
    \toprule
    Phoneme (CELEX) & Viseme\\
    \midrule
    b m p & P\\
    d n s t & T\\
    @ N g h k x & -\\
    l & L\\ 
    f v & F\\
    I i j & I\\
    E e & E\\
    a & A\\
    \& O Q o & O\\
    U Y u y & U\\
    r & R\\ 
    z & S\\
    S & G\\
	none & \# \\
  \bottomrule
\end{tabular}
\end{table}

\begin{figure*}[!ht]
\centering
\includegraphics[width=0.95\textwidth]{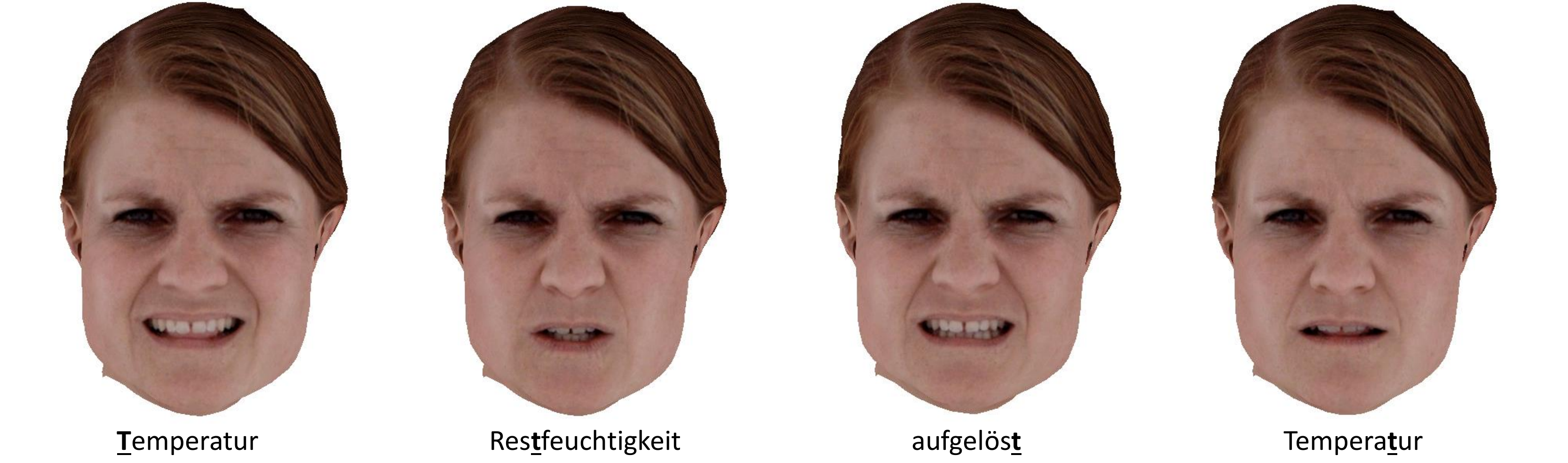}
\protect\caption{This figure shows visually different examples of German visemes for the letter 't'. }
\label{fig:coart}
\end{figure*}

An important aspect that needs to be taken into account when performing visual speech synthesis is coarticulation.

Coarticulation refers to a phenomenon in speech when two successive sounds are articulated together \cite{massaro98}, which also affects the appearance of visemes.
Figure \ref{fig:coart} shows four visually different examples of the letter 't' that are represented by the same viseme.

Therefore, we use an extended viseme label that consists of three entries for the previous, the current and the following viseme.
In addition, we introduce an 'empty' viseme marker that stands for the preceding/following viseme at the beginning and at the end of a word.
According to this notation, the German word 'kalt' (cold) is represented by the following viseme sequence: -, A, L, T and the extended viseme notation: '\#-A', '-AL', 'ALT', 'LT\#'.
In order to synthesise the mouth movement for a requested word or sentence, we create a viseme query based on $N$ extended viseme labels.
For each extended viseme $i$ in the query, we choose the optimal sample $s_i$ from our database according to the objective function (\ref{eq:VisSelection}).
\begin{equation}
\begin{split}
\mathcal{E}_{\mbox{vis}}(S)=&\sum_{i=1}^{N}\mathcal{U}(s_{i}) + \lambda\sum_{i=2}^N\mathcal{B}(s_{i}, s_{i-1})
\label{eq:VisSelection}
\end{split}
\end{equation}
The objective consists of two terms: a data term $\mathcal{U}(s_{i})$ ensures that for each viseme $i$ in the query, a suitable (i.e. according to the extended label) sample $s_i$ is chosen from the viseme database. The data term is only able to penalise deviations according to the label.
An additional neighbourhood term $\mathcal{B}(s_{i}, s_{i-1})$ minimises the differences at transitions between concatenated visemes and resolves the ambiguity that
multiple viseme samples could yield exactly the same data error $\mathcal{U}(s_{i})$.
As the optimal transition location between two viseme sequences does not depend on the viseme query, it is possible to pre-compute the optimal transition points for all combinations of two consecutive viseme samples. While this is computationally expensive, it has to be done only once for a viseme database.
We estimate the transition cost $\mathcal{B}(s_{i}, s_{i-1})$ between two viseme samples by computing the mean squared error (MSE) in latent parameter space over a temporal window. This accounts for the static appearance mismatch as well as the mismatch in dynamics. 
In order to solve for the optimal sample sequence, we interpret the objective (\ref{eq:VisSelection}) as a discrete optimization task.
For each viseme query, we have a restricted set of candidate samples that can be used to synthesise the resulting animation sequence.
This candidate set is a subset of the visme database and is created by selecting all samples of the best matching class for each viseme in the query.


Finally, we solve for the optimal viseme sequence using the alpha expansion algorithm \cite{Boykov2001}. 
Knowing the optimal viseme samples, we concatenate the parameter sequences of all selected visemes, and generate a seamless sequence of animation parameters by gradient domain blending of parameters between consecutive visemes.
After blending, we reconstruct the mouth geometry and texture with the previously trained VAE.
The reconstruction is performed frame by frame.
We integrate the synthesised mouth geometry and texture with the original 3D head model using Poisson image editing \cite{Perez2003} and Laplacian mesh editing. The resulting head model is rendered with a standard 3D graphics pipeline (e.g. OpenGL). 


\section{Experimental Results and Discussion}\label{sec:results}
This section presents still images of the proposed animation technique. We synthesised the mouthing of several German words that are not part of the viseme database.
To visualise the quality of the synthesised animation sequence, we show the selected mouth expression for each viseme and compare it to a captured version of the same expression.
Note that the results can best be evaluated in motion and we therefore also refer to the video in the supplementary material.
For our experiments, we captured an actress with 16 synchronised video camera pairs (5120x3840@25fps) that are equally placed around the actress. Four camera pairs were located on the ground, 8 camera pairs were placed at eye level and four more camera pairs were placed above the actress. We used all cameras during pose and shape estimation of the face. For the creation of dynamic face textures, we restricted the input to four camera pairs at eye level that captured the face and the left/right side of her head. Additionally, we captured several static facial expressions with 14 synchronised D-SLR cameras (Canon 550D) that were placed around the actress as well. Based on these still images and the proposed method in \cite{paier2020}, we computed an animatable head model that serves as proxy geometry in our experiments.
While we used exsting high quality hardware during data acquisition, our method is not restricted to this hardware configuration.
A low cost capture setup could for example consist of three video cameras to record the subject’s face from the left side, from the right side and frontally. 
Instead of creating a personalised face model, it is also possible to use an existing blend-shape model (e.g.~\cite{Cao2014}) or
a single 3D reconstruction of the subject’s face (e.g. neutral expression) that is deformed based on the optical flow in the video sequences (e.g.~\cite{Paier2017,Kettern15}).
This is possible as our method does not rely on accurate and detailed 3D geometry for animation.
For example, figure \ref{fig:perfcap} shows different facial expressions with a closed mouth, wide opened mouth, visible tongue and teeth.
The proxy geometry needs to account only for large scale deformations (i.e. the jaw movement).
Details like tongue or teeth are captured in texture space (see figure \ref{fig:perfcap}, left column).
All preprocessing steps, network training and experiments have been carried out on a regular desktop computer with 64GB Ram, 2.6 Ghz CPU (14 cores with hyper-threading) and one GeForce RTX2070 graphics card.
\newpage
\begin{strip}
\centering
  \includegraphics[width=0.95\textwidth]{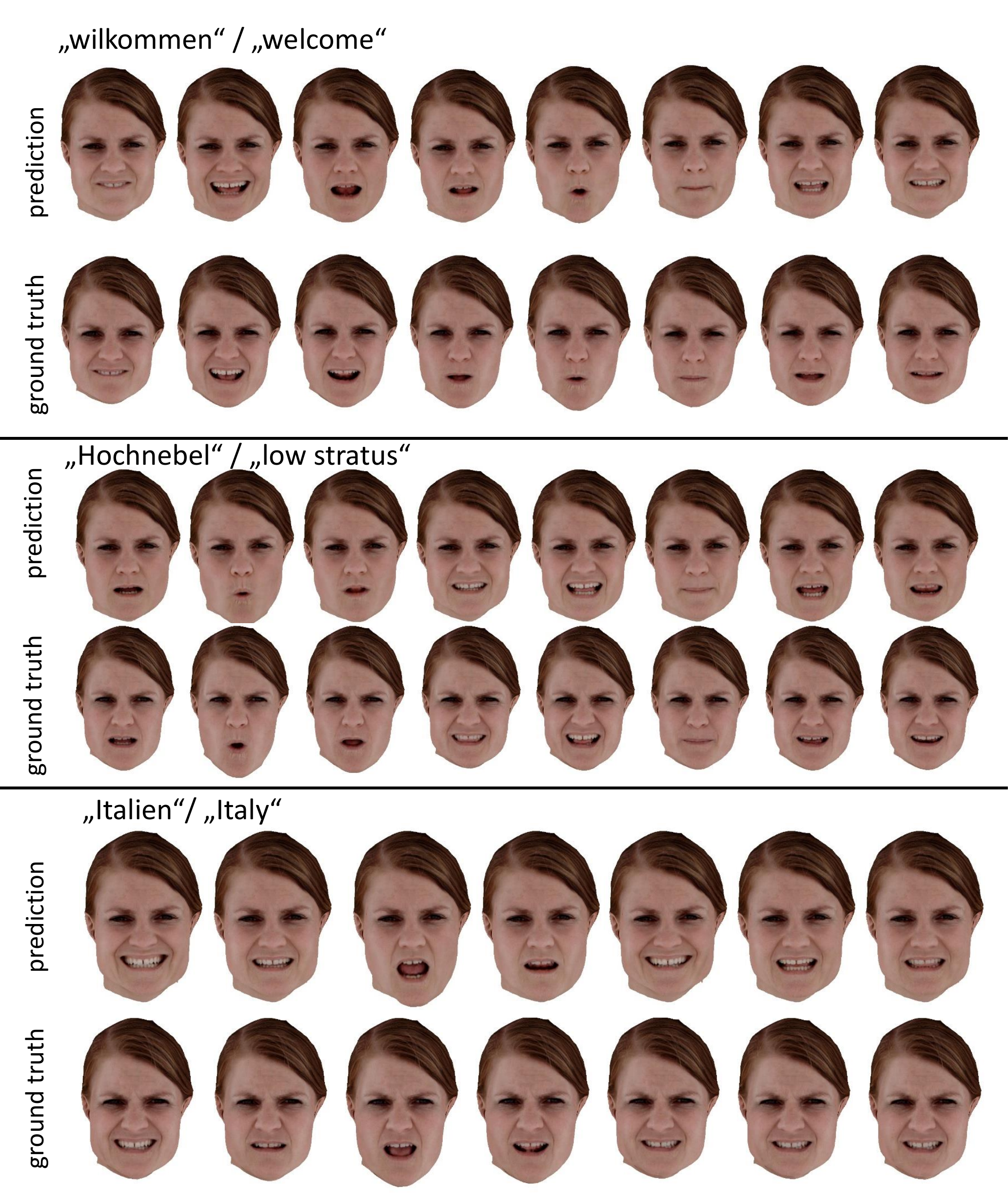}
  \captionof{figure}{This picture shows the synthesised visual speech for three words that are not part of the viseme database.}
  \label{fig:results1}
\end{strip}
\clearpage
\begin{strip}
\centering
  \includegraphics[width=0.9\textwidth]{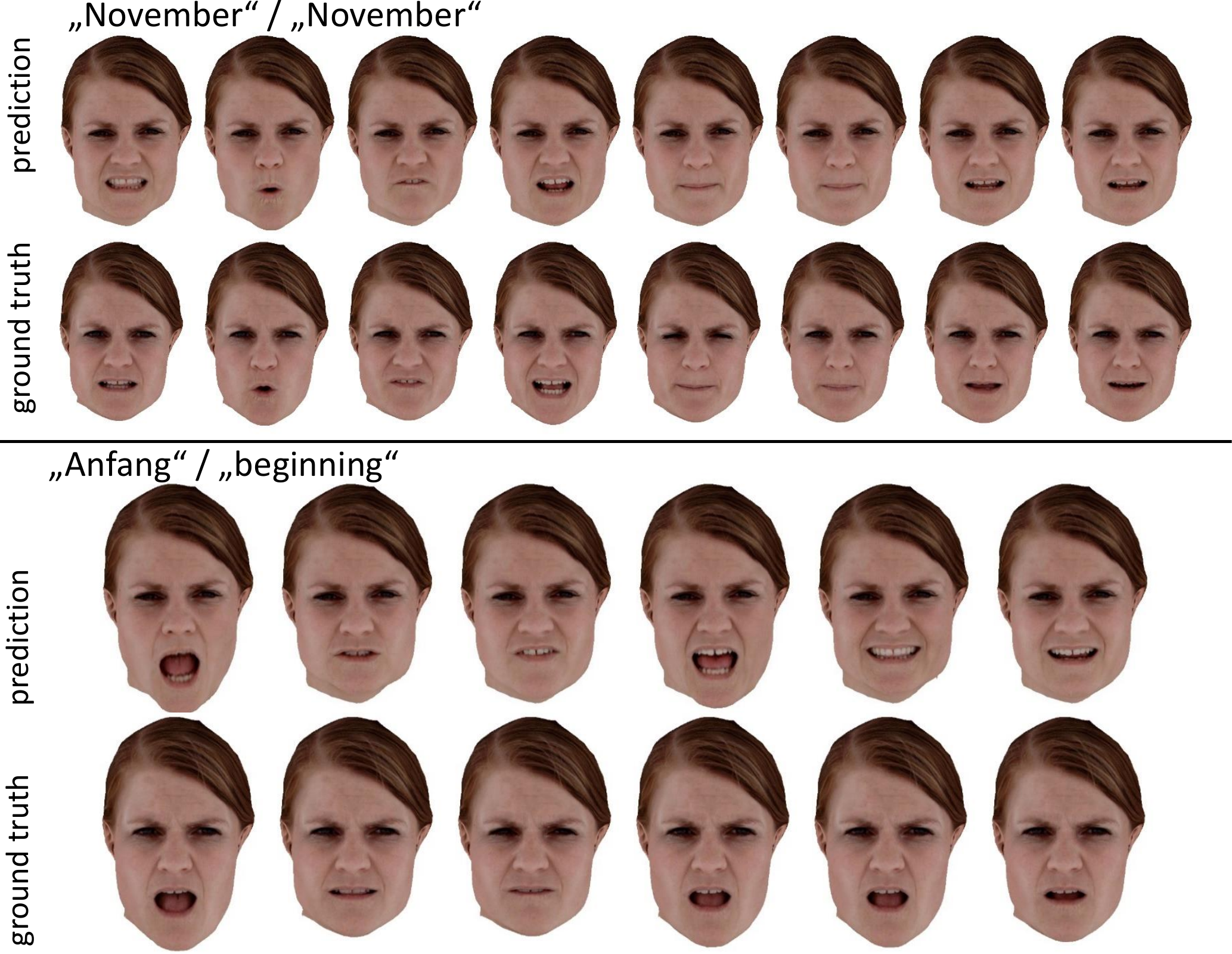}
  \captionof{figure}{This picture shows two more synthesised words that are not part of the viseme database.}
  \label{fig:results2}
\end{strip}
While the pre-processing (i.e. performance capture, network training, transition search) is done offline,
the rendering runs at a rate of approx. 40 Hz.
An unoptimized and single-threaded animation parameter synthesis takes on average 187 msec. per word.
The minimum, median and maximum computation times are 24, 151 and 578 msec. respectively.
The average query has 6 visemes. The minimum, median and maximum query lengths are 2, 5 and 16.
Figure \ref{fig:results1} and \ref{fig:results2} show that the proposed system is capable of producing correct mouth expressions according to different viseme queries.
Changes in appearance that are caused by co-articulation can be reproduced as well (compare the first expression ('I') in 'Italien' with the second expression ('i') in 'wilkommen'). 
The current neural texture model does not account for view dependency of textures.
While this is not a major limitation in most cases, it can cause erroneous perspective distortions if the geometry is not accurate
enough and the render viewpoint differs much from the camera’s viewpoint from which the texture was generated.
Another limitation of our approach are missing viseme samples in the animation database.
Due to coarticulation, the appearance of vismes can change. While it is often possible to replace a certain viseme with a similar instance of it, large deviations may reduce comprehensibility or decrease the perceived quality of the synthesised visual speech. This is shown in figure \ref{fig:results2}. For the word 'Anfang', visemes 4, 5 and 6 do not match very well. Especially viseme 5 gives the impression of speaking the phoneme n (CELEX) while the correct phoneme is N (CELEX).

\section{Conclusion}\label{sec:con}
We present a new method for example based synthesis of visual speech using a neural face representation.
Our approach is based on a high quality set of atomic/typical visual speech samples (dynamic visemes)
that are concatenated in an optimal way to produce novel sequences.
In contrast to regression-based approaches, which may suffer from under-articulation, using captured data of real people yields a richer and more precise articulation.
In order to achieve a high visual quality, we create a database of visual speech samples with a marker-less facial performance capture approach.
While using only a low detail PCA model of the actress face, we are still capable of reproducing facial expressions in high fidelity.
This is achieved by employing a novel neural face representation that captures the approximate geometry as well as a highly detailed facial texture in a compact latent vector.
In addition, we are able to solve two main problems of concatenative animation approaches: high memory requirements and the need for realistic interpolation of
facial expressions (geometry as well as texture). 

\begin{acks}
This work has partly been funded by the European Union’s Horizon 2020 research and innovation programme under grant agreement No 762021 (Content4All)
and grant agreement No 952147 (Invictus).
\end{acks}

\bibliographystyle{ACM-Reference-Format}
\bibliography{cvmp-sub-2020}


\end{document}